\documentclass{article}
\usepackage{spconf,amsmath,graphicx}

\title{ Learning Metric Graphs for Neuron Segmentation In Electron Microscopy Images}
\twoauthors
  {Kyle Luther}
	{Princeton University\\
    Department of Physics}%\\
  { H. Sebastian Seung }
	{Princeton University\\
    Department of Computer Science and Neuroscience}%\\
\begin{document}
%\ninept
%
\maketitle
\begin{abstract}
In the deep metric learning approach to image segmentation, a convolutional net densely generates feature vectors at the pixels of an image. Pairs of feature vectors are trained to be similar or different, depending on whether the corresponding pixels belong to same or different ground truth segments. To segment a new image, the feature vectors are computed and clustered. Both empirically and theoretically, it is unclear whether or when deep metric learning is superior to the more conventional approach of directly predicting an affinity graph with a convolutional net. We compare the two approaches using brain images from serial section electron microscopy images, which constitute an especially challenging example of instance segmentation. We first show that seed-based postprocessing of the feature vectors, as originally proposed, produces inferior accuracy because it is difficult for the convolutional net to predict feature vectors that remain uniform across large objects. Then we consider postprocessing by thresholding a nearest neighbor graph followed by connected components. In this case, segmentations from a ``metric graph'' turn out to be competitive or even superior to segmentations from a directly predicted affinity graph. To explain these findings theoretically, we invoke the property that the metric function satisfies the triangle inequality. Then we show with an example where this constraint suppresses noise, causing connected components to more robustly segment a metric graph than an unconstrained affinity graph.
\end{abstract}
\section{Introduction}
\label{sec:intro}

There have been several recent proposals to apply deep metric learning to image segmentation \cite{Discriminative,Semantic-Instance-Metric}. A convolutional net generates dense feature vectors at the pixels of an image. During training, pixels within the same ground truth object should be assigned vectors that are nearby in feature space, while pixels from different objects should be assigned well-separated feature vectors. To segment a new image, the feature vectors are computed by the convolutional net, and then clustered. (The idea was also part of a more complex system for detecting and correcting segmentation errors \cite{Error-Detection}). 

\begin{figure}[htb]

\begin{minipage}[b]{1.0\linewidth}
  \centering
  \centerline{\includegraphics[width=8.5cm]{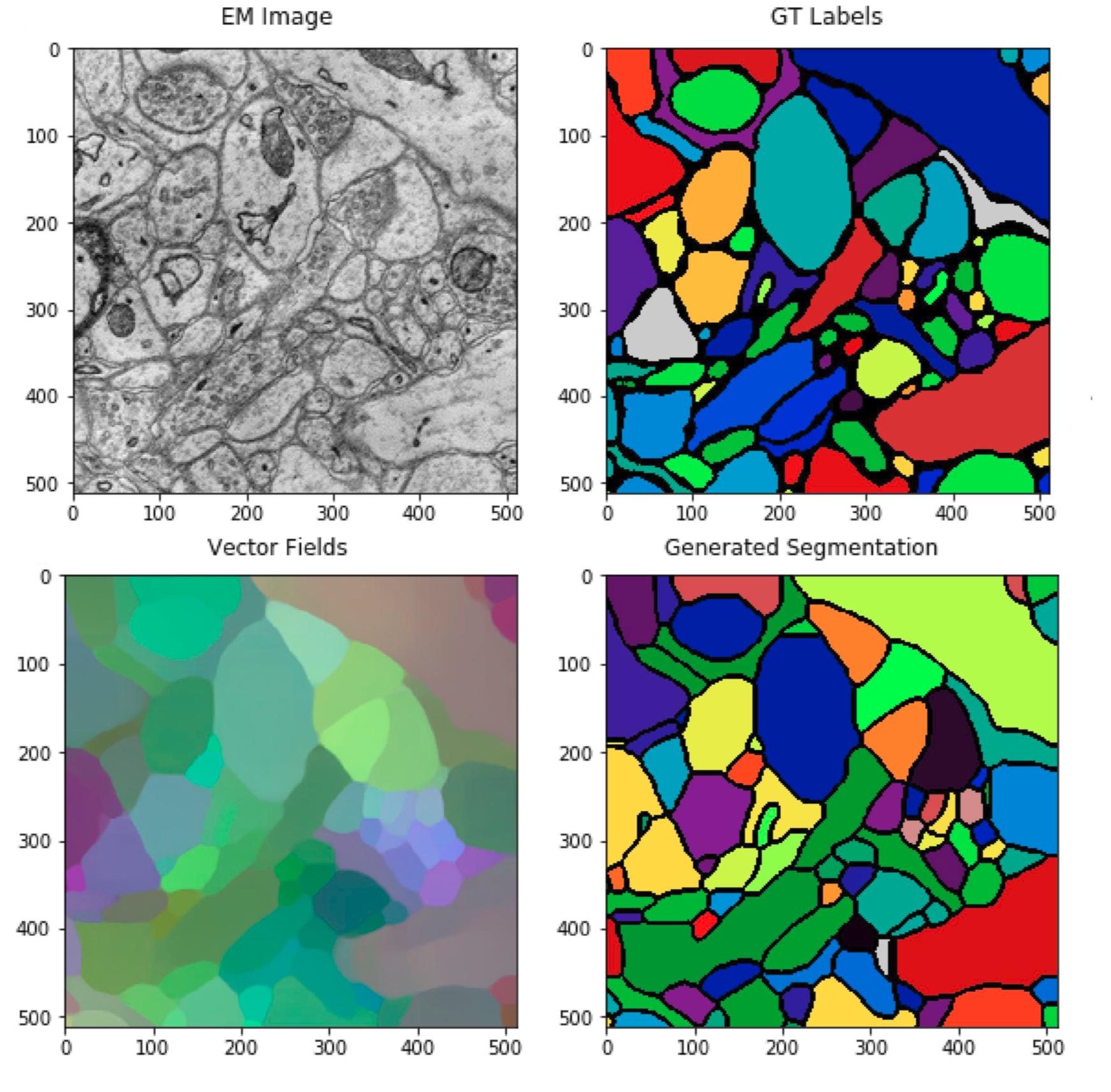}}
%  \vspace{2.0cm}
%  \centerline{(a) Result 1}\medskip
\end{minipage}
\caption{ Test set example: Upper left: crop of input EM image to network. Upper right: ground truth labels. Lower Left: PCA-based visualization of vector fields. Lower right: connected components segmentation derived from vector fields. }
\label{overview}
\end{figure}

The deep metric learning approach is conceptually intriguing because it is a hybrid of semantic and instance segmentation. Feature vectors can potentially encode semantic information about the underlying objects. At the same time, multiple instances of the same kind of object should be assigned distinct feature vectors. Deep metric learning has been applied to natural images \cite{Discriminative,Semantic-Instance-Metric}, but this is arguably not a difficult test of instance segmentation. Many natural images contain only a few objects, or a few instances of each object, and instances are often well-separated in the image. Furthermore, many instances of the same object class have very different "low-level" features (shape, color, texture, etc.), and therefore are not hard to differentiate "semantically." 

As a more challenging test of instance segmentation, we use neuronal images acquired by serial section electron microscopy (EM). An image of $400\times 400$ pixels can contain a rather large number of objects (more than 50 neuronal cross sections). Furthermore, many objects have similar low-level features (intensity, texture, shape) and assigning them distinct feature vectors would seem to require encoding contextual information about surrounding objects. A state-of-the-art approach to this problem is to train a convolutional network to directly predict affinities between nearest neighbor pixels, essentially identifying the boundaries between objects, and then partition the affinity graph \cite{turaga2010convolutional,Superhuman}. 

We start by showing that seed-based postprocessing schemes proposed by \cite{Discriminative, Semantic-Instance-Metric} yield low quality segmentations on this task, in large part because the vector fields do not remain uniform over large distances. We then show that using a simpler approach, connected components, on the vector fields yields segmentations of far higher quality, surpassing even the scores of connected components on directly predicted affinity graphs, which are the backbone of many state-of-the-art segmentation pipelines \cite{turaga2010convolutional,Superhuman}.

In the final section we provide an possible reason for the improved performance of the vector-derived segmentations over the affinity-derived segmentations. We show that connected components on the vector fields is equivalent to connected components on an affinity graph where the affinity graph is constrained to satisfy the triangle inequality (called a "metric graph"). We then show an example where this constraint suppresses errors in an unconstrained affinity graph, suggesting that connected components may be more robust to effective on a metric graph than an unconstrained affinity graph.

\section{Methods}

%\subsection{Metric Graphs}
We call a metric graph any weighted graph whose weights satisfy the axioms of a metric. In our application the nodes of the graph are pixels and the edges (typically nearest neighbor) between pixels. We can derive an affinity graph from a metric graph by simply inverting the signs of all the distances between objects (i.e. affinity is the negative distance between nodes).
We note that general affinity graphs however cannot always be derived from a metric graph. In particular, there is no requirement that the affinities be consistent with the triangle inequality.  We show later that the additional requirement that the edge weights be derived from a metric gives us some theoretical properties that seem desirable for generating a segmentation.

We use the concept of a metric graph to encode the notion of locality. In particular, we will show that a network can learn distances over short range edges much more accurately than over long range edges. Using just the short range edges allows us to generate more accurate segmentations than other seed-based approaches

We represent our metric graphs using a vector at each pixel and compute the edge weight between a pair of pixels using the L1 norm between vectors. This ensures the edge weight satisfies the triangle inequality.

\subsection{Means-based loss function}
We use convolutional networks to generate the vector fields. We use the loss presented by \cite{Discriminative} to train our networks. Briefly, there are three terms:
$ L_{int} = \frac{1}{C} \sum_{c} \frac{1}{N_c} \sum_{i} \Vert \mu_c - v_i \Vert^2 $, 
$ L_{ext} = \frac{1}{C(C-1)} \sum_{c_a, c_b:c_a \neq c_b} \max(2 \delta_d - \Vert \mu_{c_a} - \mu_{c_b} \Vert, 0)^2 $, $L_{norm} = \frac{1}{C} \sum_c \Vert \mu_c \Vert$, 
where $v_i$ is the vector for pixel i, $\mu_c$ is the mean vector for object c, C is the number of objects, $N_c$ is the number of pixels in object c, $\Vert \cdot \Vert$ is the L1 norm, $\delta_d$ is the margin of the external loss (we only care if vectors are further apart than some threshold). We compute the total loss as:
\[L = L_{int} + L_{ext} + \gamma L_{norm} \]
We use an embedding dimension of 32, $\delta_d=1.5$ and $\gamma=0.001$. The authors in \cite{Discriminative} also had a margin in $L_{int}$ but we found that removing this margin yielded smoother vector fields within objects and did not hurt the networks ability to minimize $L_{ext}$.

%We note that this dataset presents an extra challenge not discussed in \cite{Discriminative, Semantic-Instance-Metric}. There may be two parts of an object that appear to be separate objects in a given patch, but there exists a path connecting the two parts outside the patch under consideration. This situation presents itself relatively frequently and training the network to assign similar vectors to these two object parts was found to hurt performance. We find that splitting objects in each training patch if there is no path between them and then removing all terms in $L_{ext}$ between the newly formed split objects yields better generalization performance than either training the network to assign vectors to different object parts or ignoring the complication all together.

We observe that there may be two parts of an object that appear to be separate objects in a given patch, but there exists a path connecting the two parts outside the patch under consideration. We find that splitting objects in each training patch if there is no path between them and then removing all terms in $L_{ext}$ between the newly formed split objects yields better generalization performance than either training the network to assign vectors to different object parts or ignoring the complication all together.

%We also note that there are some pixels (in particular, pixels that are on or outside cell boundaries) that are labeled as background pixels. We do not compute the loss on these pixels.

\subsection{Network architecture}
We use a variant of the popular U-Net \cite{UNet} architecture. 
It is a fully convolutional network consisting of a downsampling followed by upsampling path and uses skip connections along the path. We modify the original architecture in 3 ways. One, we use batch normalization \cite{BatchNorm} before each nonlinearity. We use the dynamic variant of batch norm meaning at inference time, we use the layer statistics for the particular image being processed rather than the statistics averaged over the training set. Two, we add one more level of resolution to the network so our network consists of a total of 6 levels of resolutions. This additionally increases the receptive field of our network to approximately 400x400 pixels. Finally, we add additional skip connections within the convolutional blocks at each resolution as done in \cite{Superhuman}.

\subsection{Dataset}
We use EM brain images from the AC3 and AC4 datasets \cite{Saturated}. We use a 1024x1024x256 stack for training and a 1024x1024x100 for testing.  The labels are densely labeled integer ids corresponding to the segmentation of the images. The images are a superset of the popular SNEMI3D reconstruction challenge where contestants attempted to generated 3D neuron reconstructions \footnote{SNEMI3D is similar to the popular 2012 ISBI neuronal segmentation challenge. We use this primarily because it is larger dataset. }. Here we conduct all experiments in 2D as it is simpler and our results should not be fundamentally tied to the dimensionality of the dataset. An image and label are shown in Figure~\ref{overview}.

\subsection{Training details}
%pixel pairs from foreground objects
We sample a crop of size 924x924 from 1 section and compute the loss as described above. We use the Adam optimizer \cite{Adam} with a learning rate of 0.001. We train each network for about 50K iterations, taking approximately 20 hours on a single NVIDIA Titan X Pascal GPU.

For comparison with state of the art methods, we additionally train a second network using the same procedure/architecture/augmentation but instead train it to directly produce affinities between nearest neighbor pixels using the cross entropy loss. \cite{turaga2010convolutional, Superhuman}.

We apply standard augmentation procedures: random rotations, flips, and rescaling. Additionally we apply elastic deformation augmentation as described in \cite{UNet}. This applies a low frequency Gaussian displacement field to the images and labels, artificially enlarging the training set. 

\subsection{ Visualization }
\begin{figure}[htb]

\begin{minipage}[b]{1.0\linewidth}
  \centering
  \centerline{\includegraphics[width=8.5cm]{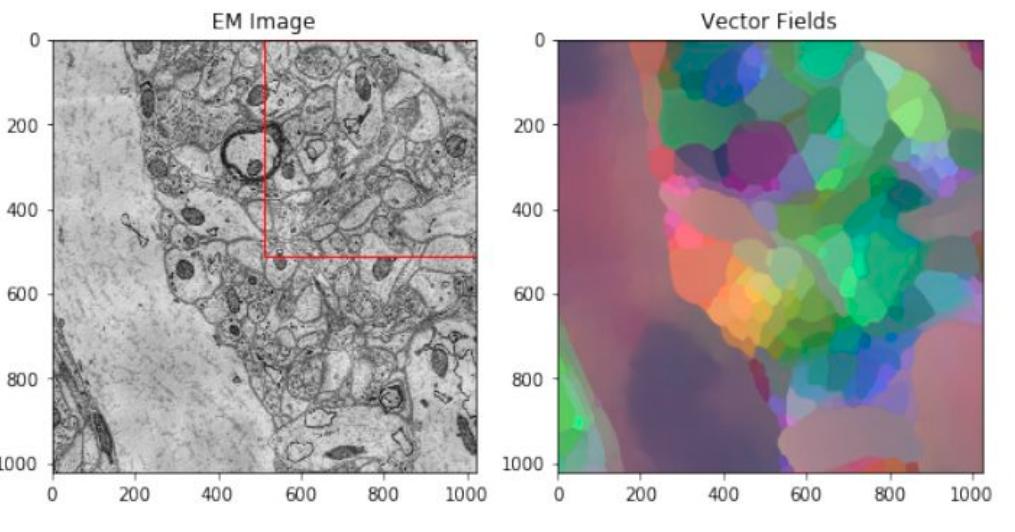}}
%  \vspace{2.0cm}
%  \centerline{(a) Result 1}\medskip
\end{minipage}
\caption{ Left: Input EM image (red box contains section in Figure~\ref{overview}). Right: PCA  visualization of vector fields. Note the nonuniformity in the vector fields in the largest neuron. }
\label{nonuniform}
\end{figure}

For visualization, we  project the vector fields onto the top 3 components given by principle components analysis (PCA) \cite{PCA} and visualize this projection as an RGB image. Examples are shown in Figure~\ref{overview} and Figure~\ref{nonuniform}. A striking feature of this visualization is that these projections look like segmentations that would normally be obtained after some form of post-processing. However, these are visualizations of the direct output given by the CNN, with no additional post-processing besides the PCA projection. 

\section{Segmentations}
We describe the segmentation and evaluation procedures below. An example segmentation generated using connected components on the vectors is shown in Figure~\ref{overview}.

\begin{figure}[htb]

\begin{minipage}[b]{1.0\linewidth}
  \centering
  \centerline{\includegraphics[width=7.5cm]{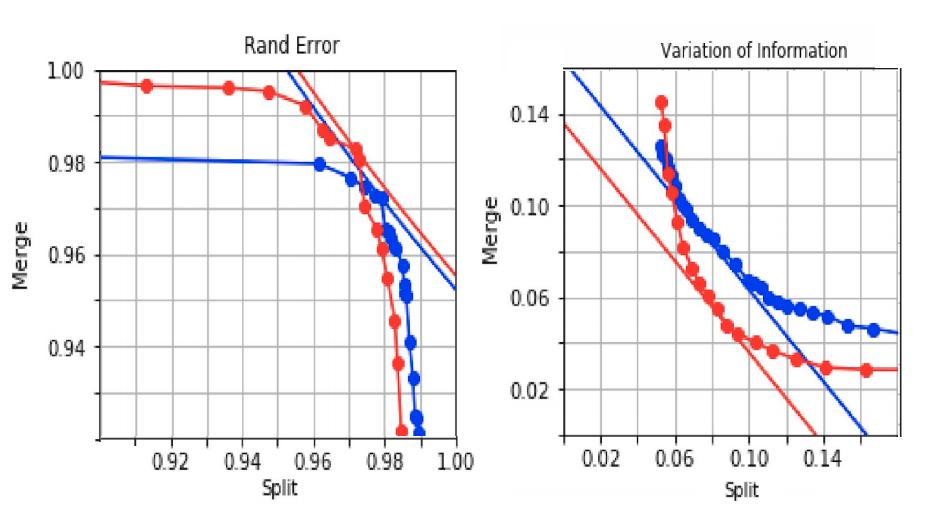}}
%  \vspace{2.0cm}
%  \centerline{(a) Result 1}\medskip
\end{minipage}
\caption{Segmentation Scores. Left: Rand F-Score. Right: Variation of Information. Segmentation of the metric graphs seems to yield significantly fewer merge errors.}
\label{scores}
\end{figure}

\subsection{Postprocessing}
We partition our affinity graph by running connected components on the nearest neighbor edges in the graph \cite{CC}. This generates reasonable segmentations but tends to produce lots of singleton labels near the objects of borders. To make evaluation fair between all methods, we label all singleton objects as background and dilate all segments by up to 10 pixels to reduce the amount of unlabeled pixels.

For the seed-based segmentation, we use the ground truth object masks to generate mean vectors and simply connect each pixel to the nearest mean vector. This should put a rough upper bound on seed-based segmentation performance (either for seeds predicted from a network \cite{Semantic-Instance-Metric} or seeds given by MeanShift \cite{Discriminative}). It was also empirically demonstrated \cite{Discriminative} that this method performed better than MeanShift which does not have access to ground truth labels.

\subsection{Evaluation}
We quantitatively evaluate the segmentations through two metrics, the Rand F-Score, R, and the Variation of Information, V, two popular metrics used to evaluate segmentation of connectomic images \cite{Crowdsourcing}. Both metrics can be split into merge and split scores which measure the degree to which one splits and mergers affect the segmentation performance.  We do not evaluate any quantity on any pixel within 2 pixels of a boundary in the ground truth. This is done to reduce the sensitivity to the precise locations of the boundaries. 

Figure~\ref{scores} shows the scores of the connected component segmentations on the affinity graphs. We observe that by both metrics, the vector fields seem to provide better connected components segmentations. We note that the improvement in the merge scores are relatively larger than the improvement in the split scores. As we argue in the following section, this may be a result of the restriction on the affinity graphs that is imposed by requiring the affinities be derived from a metric graph.

Additionally we find that the scores for the seed-based segmentation are far lower than the connected components segmentations and they do not show up on the graphs. In particular, the seed-based segmentation receives a Rand Score of 0.71 and a VI Score of 0.59. This is largely due to the fact that the vector fields are nonuniform within large objects (see Figure~\ref{nonuniform}), and this causes many split segments.

\section{Connected Components on a Metric Graph}

%In this section we compare affinity graphs derived from metric graphs and affinity graphs that do not have any constraints on the affinities.

%\subsection{Triangle Inequality Imposes Constraints on the Affinity Graph}
By definition, any triplet of nodes $i,j,k$ in a metric graph must satisfy $d_{ik} \leq d_{ij} + d_{jk}$. This implies that not all affinity graphs can be derived from a metric graph. However, for the purposes of segmentation, we care primarily about a particular class of affinity graph: affinity graphs where the distance between nodes can be written as $d_{ij} = 1-\delta_{ij}$. Here $\delta_{ij}$ is the Kronecker delta function and is 1 when $i$ and $j$ are given the same label and 0 otherwise. Clearly this is a metric (it is non-negative, symmetric, satisfies triangle inequality, and 0 iff $i$ and $j$ have the same label). Therefore an affinity graph derived from a segmentation can always be represented by a metric graph.

\subsection{Approximating Affinity Graphs with Metric Graphs}

\begin{figure}[htb]

\begin{minipage}[b]{1.0\linewidth}
  \centering
  \centerline{\includegraphics[width=7.5cm]{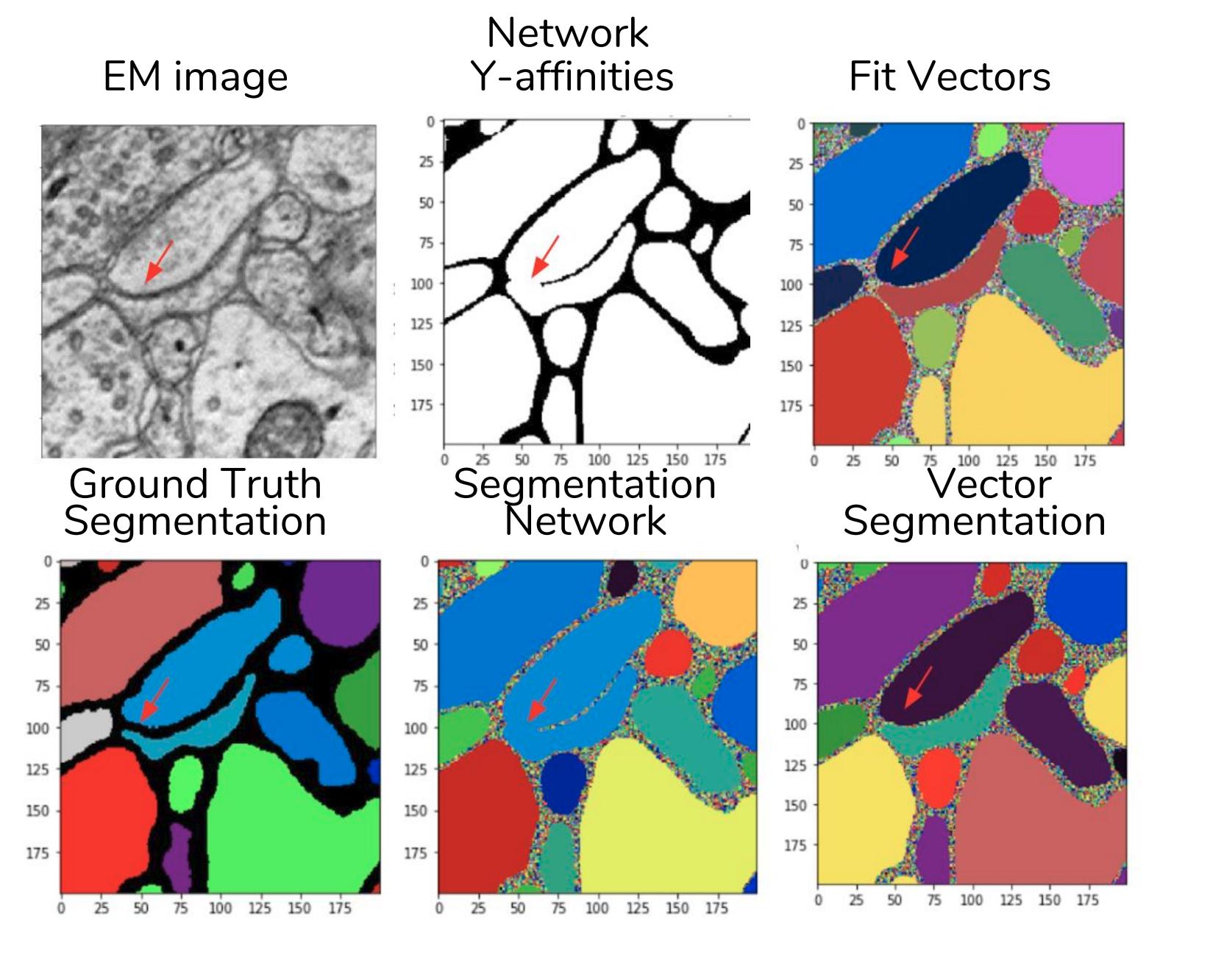}}
%  \vspace{2.0cm}
%  \centerline{(a) Result 1}\medskip
\end{minipage}
\caption{Example from the test set where requiring the predicted affinity graph to satisfy the triangle inequality can improve the resulting segmentation. }
\label{example_error}
\end{figure}

Here we provide anexample which suggests that requiring a learned affinity graph to satisfy the triangle inequality can lead to improvements in the derived segmentation. Essentially, we find a location where the affinity graph output by our baseline network contains noise (not that it is not uncorrelated "shot noise", rather it is spatially distributed over many pixels). This noisy affinity graph is inconsistent with an underlying metric. Finding an approximating metric graph is enough to remove this noise and fix an error in the resulting segmentation. The results are shown in Figure~\ref{example_error}. 

More precisely, we trained a network to directly generate affinities between sparsely sampled pixel pairs up to 32 pixels apart. This "long-range affinity prediction" is described in \cite{Superhuman}. This procedure is identical to the procedure used to train our baseline direct affinity network except that it outputs predictions over many pixel pairs. We only show the nearest neighbor affinities in the vertical direction in the figure.

We then generate affinity graphs for all images in the test set. We then find the connected components of the thresholded nearest neighbor affinity graph. Two objects from the test set are erroneously merged (red arrow) as shown in Figure~\ref{example_error}. Now we define a 3-element embedding vector at each pixel and define the affinity between two vectors as $e^{-d}$ where $d$ is the distance. We optimize the vectors to minimize the squared error relative to the directly generated affinity graph. Note that these vectors are not learned, rather they are derived directly from the affinity graph. Taking only the nearest neighbor edges of this regularized affinity graph, we threshold and compute connected components, and the merge error is eliminated. This example is anecdotal, but it does illustrate the principle that merely representing affinities using embedding vectors can clean up errors in an affinity graph, even if those embedding vectors are not generated by a neural net.

Intuitively what happened in this example was the network erroneously predicted "merge" affinities on the left side of the merged objects and correctly predicted "split" affinities on the right side. Additionally, the network correctly predicted "merge" affinities between the left and right sides of each object. This resulted in an \emph{inconsistent} affinity graph. Fitting vectors to this affinity graph in the manner prescribed required that the vector-based affinities were consistent with the triangle inequality and doing this required some sort of averaging of predictions to fit the vectors.

This research was supported by the Intelligence Advanced Research Projects Activity (IARPA) via Department of Interior/ Interior Business Center (DoI/IBC) contract number D16PC0005. The U.S. Government is authorized to reproduce and distribute reprints for Governmental purposes notwithstanding any copyright annotation thereon. Disclaimer: The views and conclusions contained herein are those of the authors and should not be interpreted as necessarily representing the official policies or endorsements, either expressed or implied, of IARPA, DoI/IBC, or the U.S. Government.

%This method could be of practical use in modern connectomics pipelines. Having a simple post-processing scheme is likely important for large scale neural reconstructions. The downside of such simplicitiy is often a reduction in accuracy. We have provided a method that improves the accuracy of the simple connected components scheme while adding virtually no inference time overhead.

\bibliographystyle{IEEEbib}
\bibliography{strings,refs}

\end{document}